\documentclass[letterpaper,10pt, conference]{ieeeconf}

\IEEEoverridecommandlockouts 

\usepackage{amsmath,amsfonts}
\usepackage{array}
\usepackage{amssymb}
\usepackage{algorithmic}
\usepackage{algorithm}
\usepackage{array}
\usepackage{bm}
\usepackage{xcolor}
\usepackage{float}
\usepackage{threeparttable}
\usepackage[caption=false,font=normalsize,labelfont=sf,textfont=sf]{subfig}
\usepackage{textcomp}
\usepackage{stfloats}
\usepackage{url}
\usepackage{booktabs}
\usepackage{multirow}
\usepackage{lettrine}
\usepackage{verbatim}
\usepackage{tabularx}
\usepackage{graphicx}
\usepackage{cite}
\usepackage[colorlinks=true, linkcolor=red, citecolor=red, urlcolor=red]{hyperref}
\begin{document}

\title{\LARGE \bf
Interpretable Interaction Modeling for Trajectory Prediction via Agent Selection and Physical Coefficient}

\author{Shiji Huang, Lei Ye*, Min Chen, Wenhai Luo, Dihong Wang, Chenqi Xu, Deyuan Liang
\thanks{S. Huang, L. Ye, M. Chen, W. Luo, D. Wang, C. Xu, D. Liang are with the College of Computer Science and Technology, Zhejiang University of Technology, China (email: \{shiji, yelei, cm, whailuo, wangdh, xucq, liangdy\}@zjut.edu.cn). *Corresponding author: Lei Ye.}}



\maketitle

\begin{abstract}
A thorough understanding of the interaction between the target agent and surrounding agents is a prerequisite for accurate trajectory prediction. Although many methods have been explored, they assign correlation coefficients to surrounding agents in a purely learning-based manner. In this study, we present ASPILin, which manually selects interacting agents and replaces the attention scores in Transformer with a newly computed physical correlation coefficient, enhancing the interpretability of interaction modeling. Surprisingly, these simple modifications can significantly improve prediction performance and substantially reduce computational costs. We intentionally simplified our model in other aspects, such as map encoding. Remarkably, experiments conducted on the INTERACTION, highD, and CitySim datasets demonstrate that our method is efficient and straightforward, outperforming other state-of-the-art methods. 
\end{abstract}


\section{Introduction}
The ability to accurately forecast the trajectories of human-driven vehicles and pedestrians sharing the environment with autonomous vehicles is paramount within autonomous driving. Such precise trajectory predictions are indispensable for downstream intelligent planning systems to make informed decisions, thereby improving autonomous driving operations' safety, comfort, and efficiency. However, due to the inherent uncertainty and the multimodal nature of driving behaviors, vehicle trajectory prediction against an urban setting presents significant challenges that include, but are not limited to, spatio-temporal modeling of historical trajectories\cite{Janjos2021StarNetJA}, interaction modeling\cite{Mo2022MultiAgentTP,Scibior2021ImaginingTR}, environmental description\cite{Gao2020VectorNetEH}, kinematic constraints\cite{Scibior2021ImaginingTR,Janjos2021SelfSupervisedAP}, and real-time inference\cite{Mao2023LeapfrogDM}.

\begin{figure}[!t]
\centering
\includegraphics[width=1.0\columnwidth]{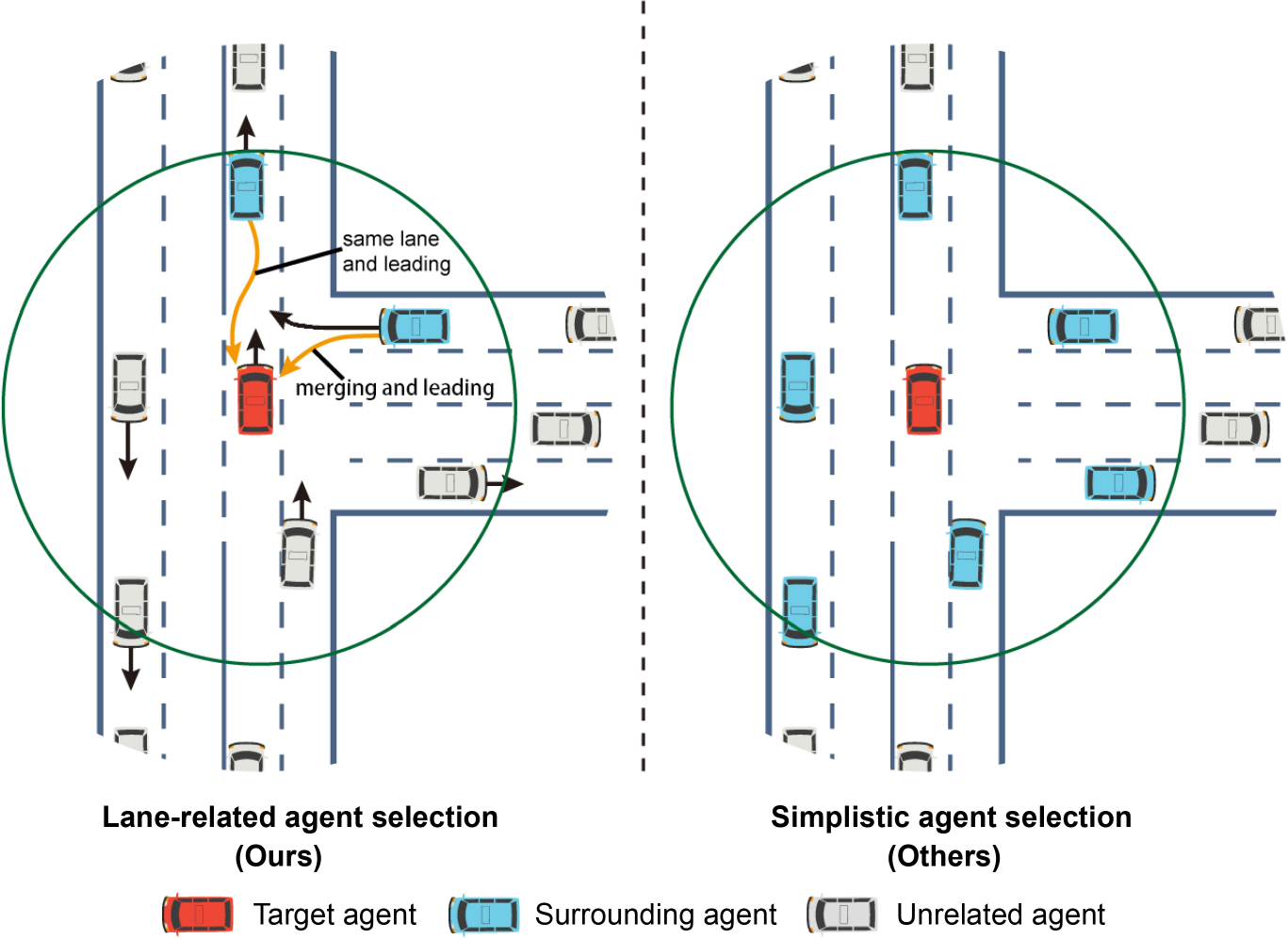}
\caption{Left: We first predict each agent's future lane. Then, combine it with their current lane and direction relative to the target agent to select interacting agents further. Right: Classical agent selection method, which takes all agents within a defined range as model input.}
\label{fig1}
\end{figure}

Recent studies\cite{Zhou2022HiVTHV,Mo2022MultiAgentTP,Scibior2021ImaginingTR,Shi2023MTRMM,Zhou2023QueryCentricTP} focus on modeling interactions between agents, as it is a crucial element in autonomous driving. For example, HiVT\cite{Zhou2022HiVTHV} and QCNet\cite{Zhou2023QueryCentricTP} leverage the attention mechanism to model agent-agent interactions, which can implicitly select significant nearby agents for the target agent. The larger attention weights indicate the importance of these agents. Although the results of these studies demonstrate the superiority of their interaction modeling, they do not elucidate the underlying decision logic or cognitive processes. To this end, we enhance the interpretability of interaction modeling from the following two aspects:

(\romannumeral1) {\textit{Agent Selection}}. Previous methods tend to take all surrounding agents as input to the interaction module. However, human attention capacity is limited. In dynamic environments, one person can focus on at most five agents at a time\cite{Pylyshyn1988TrackingMI}, which means that a huge amount of irrelevant agents are also input into the model. As depicted in Fig.~\ref{fig1}, we refine the selection of interacting agents by leveraging their road topology, thus improving explainability. More importantly, this enables the model to swiftly filter out unrelated agents, markedly decreasing inference latency in interaction-rich environments.

(\romannumeral2) {\textit{Interaction Encoding}}. The attention mechanism\cite{Janjos2021StarNetJA} and the graph neural network (GNN)\cite{Knittel2022DiPAPM} are popular in numerous interaction-aware methods. In this work, we quantify the correlations between agents using a handcrafted simple physical attention score and replace the traditional attention score by integrating it into the Transformer framework. Specifically, the physical attention score is obtained by normalizing the closeness index between agents, where the closeness index accounts for both the distance and the speed of approach.

We incorporate these two improvements into a simple Conditional
Variational Autoencoder (CVAE) framework, named ASPILin, which first predicts the Gaussian distribution parameters of future trajectories and then generates multimodal trajectories through reparameterization.
Additionally, in contrast to the majority of existing approaches\cite{Gilles2021GOHOMEGH, Sun2022M2IFF,Knittel2022DiPAPM}, we conduct agent selection for each historical time step (rather than only the current time). Consider a scenario following an overtaking maneuver, where at the current time, vehicle A has already overtaken and is positioned in front of vehicle B. Our agent selection approach does not regard vehicle B as an interacting agent in this situation. However, in the historical time steps, A considered B to be an interacting agent. Within the framework of our agent selection approach, exclusively selecting potential interacting agents at the current timestamp may lead the model to disregard the interaction behaviors of the target agent in historical timestamps. Certainly, for other models, conducting agent selection solely at the current time is entirely reasonable, as their simplistic agent selection methods diminish the variability of selection across different time steps. We deliberately use simple network structures to simplify our model for the other two modules (historical trajectory encoding and map encoding). Comparative experiments on the INTERACTION\cite{Zhan2019INTERACTIONDA} and highD\cite{Krajewski2018TheHD} datasets demonstrate that ASPILin is highly competitive with other state-of-the-art methods. More importantly, ablation studies on the INTERACTION and CitySim\cite{Zheng2022CitySimAD} datasets indicate that our improvements to the interaction module achieved better prediction performance with lower inference latency. We also highlight that these improvement strategies can be easily incorporated into other models, particularly the agent
selection strategy.

In summary, our contributions are:
\begin{itemize}
\item A heuristic agent selection method that further selects interacting agents based on the road topology of the agents.
\item A hand-crafted attention scores based on physical characterization instead of learned attention.
\item We propose a lightweight trajectory prediction model called ASPILin, which achieves competitive results on popular public datasets.
\end{itemize}

\section{Related Work}
\subsection{Interaction-Aware Trajectory Prediction}
Existing interaction-aware methods can be further explored from the perspectives of agent selection and interaction encoding.

Many methods\cite{Ngiam2022SceneTA,Liu2023LAformerTP} directly model interactions with all agents within the scene and simultaneously predict the trajectories of multiple target agents. By contrast, setting a range threshold\cite{Gao2020VectorNetEH,Scibior2021ImaginingTR,Gilles2021GOHOMEGH,Janjos2021StarNetJA,Mo2022MultiAgentTP,Knittel2022DiPAPM,Bhattacharyya2022SSLLanesSL} or limiting the maximum number of neighbors\cite{Shi2023MTRMM} permits modeling of the target agent's local context, which aligns more closely with the needs of single-agent prediction\cite{Gao2020VectorNetEH,Gilles2021GOHOMEGH,Janjos2021StarNetJA}. Moreover, many works\cite{Janjos2021StarNetJA,Shi2023MTRMM,Zhou2022HiVTHV} conduct joint predictions by initially capturing local interactions before modeling global interactions, enabling the model to extend from single-agent prediction to multi-agent prediction. Most of these methods only model the interacting agents of the target agent at the current moment. For multi-step prediction methods\cite{Scibior2021ImaginingTR} and vectorized representation methods\cite{Gao2020VectorNetEH,Zhou2022HiVTHV}, they must model the interactions for each historical timestep.
  
For interaction encoding, most studies use purely learning-based approaches. However, in other respects, recent studies combine physics- and learning-based approaches, offering insights into improving model performance. SSP-ASP\cite{Janjos2021SelfSupervisedAP} and ITRA\cite{Scibior2021ImaginingTR} limit motion learning to an action space grounded in acceleration and steering angles, subsequently deducing future trajectories via a kinematic model. M2I\cite{Sun2022M2IFF} classifies a pair of agents as influencer and reactor by calculating the closest value of their ground-truth trajectories and the time required to reach the nearest point at the training stage, followed by the sequential generation of their future trajectories via a marginal predictor and a conditional predictor, respectively.

\subsection{Multi-modal Trajectory Prediction}
The future trajectory of vehicles inherently exhibits multimodality, given the uncertainty of intentions. To tackle this challenge, one widely used approach involves modeling the output as a probability distribution of future trajectories via regression\cite{Zhou2022HiVTHV}. Usually, it introduces a cross-entropy loss function for mode classification to avoid mode collapse. Some methods use more explicit classification representations to make multi-modal trajectories closer to reality. TNT\cite{Zhao2020TNTTT} samples anchor points from the roadmap and then generates trajectories based on these anchors. SSL-Lanes\cite{Bhattacharyya2022SSLLanesSL} classifies the maneuvers of each agent and trains the model in a self-supervised manner.

Other methods parameterize the distribution of future trajectories\cite{Varadarajan2021MultiPathEI}, such as Gaussian Mixture Models (GMM) or samples within a latent space, and generate predictions through mapping. Regarding the latter, Generative Adversarial Networks (GANs)\cite{8578338}, Conditional Variational Autoencoders (CVAEs)\cite{Janjos2021StarNetJA,Janjos2023ConditionalUA}, and diffusion model\cite{Mao2023LeapfrogDM} are the most popular models. A common drawback of generative models is the need for extensive data to support training. Moreover, for GANs, challenges such as training difficulties and mode collapse exist. For the diffusion model, multi-step denoising leads to significant computational overhead and high inference latency. Although CVAEs face issues of insufficient diversity like GANs, their training is more stable. In this study, instead of sampling randomly from a standard normal distribution, we treat the sampler as a trainable module to prevent unrealistic trajectory outputs due to randomness.

\begin{figure*}[!t]
\centering
\includegraphics[width=\textwidth]{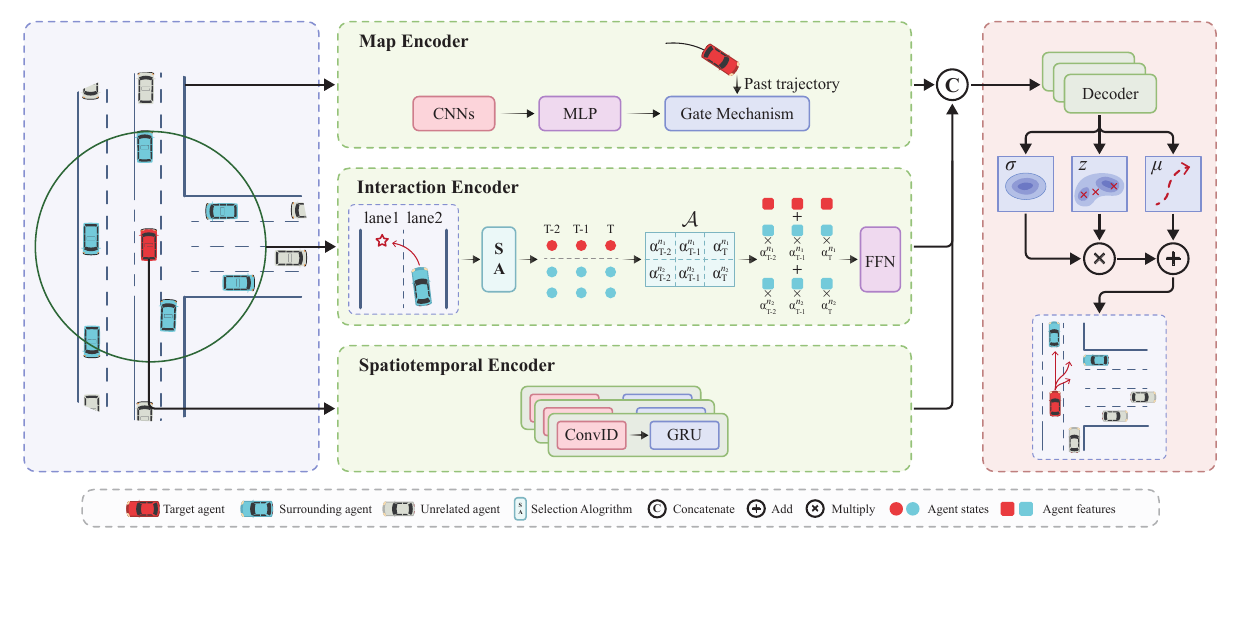}
\caption{The illustration of ASPILin. We focus on interaction modeling and deliberately simplify the design of other modules to prove the effectiveness of our method. A lane predictor and a new algorithm are used to select interacting agents further, and a novel physical correlation coefficient is designed to replace data-driven attention encoding. The multimodal trajectory predicion results are finally derived from a reparameterization formula.}
\label{fig2}
\end{figure*}

\section{Methodology}

\subsection{Problem Formulation}
Single-agent trajectory prediction is designed to forecast the future trajectory of a target agent conditioned on agents' historical states $X$ and the map information $\mathcal{M}$. To be more specific, we assume that at time $t$, there are $N$ agents (vehicles, pedestrians, cyclists) in the scene, so their historical states can be represented as ${X_{t}} = [x_{t}^{0}, x_t^{1}, \ldots, x_t^{N-1}] \in \mathbb{R}^{N \times 7}$, where $x_t^{n}$ is the state of agent $n$ at time $t$, including $xy$ coordinates ($p_t^{n} \in \mathbb{R}^2$), heading angle ($h_t^{n} \in \mathbb{R}$), speed ($v_t^{n} \in \mathbb{R}^2$), and acceleration ($a_t^{n} \in \mathbb{R}^2$). Specifically, $x_{t}^{0}$ represents the state of the target agent at time $t$. Taking into account $T_h$ historical observation timesteps, the overall historical state of the agents is denoted as $X = [X_{-T_h+1}, X_{-T_h+2}, \ldots, X_{0}] \in \mathbb{R}^{N \times T_h \times 7}$. Similarly, the future ground-truth trajectory of the target agent is defined as $Y = [y_1,y_2, \ldots, y_{T_f}] \in \mathbb{R}^{T_f \times 2}$ over $T_f$ timesteps, where $y_t$ is the $xy$ coordinates at time $t$. To forecast multi-modal future trajectories, the predicted trajectory of $K$ modes is denoted as $\hat{Y} = [\hat{Y}^{1},\hat{Y}^{2},\ldots,\hat{Y}^{K}] \in \mathbb{R}^{K \times T_f \times 2}$. Our goal is to learn a generative model to parameterize the distribution $\mathcal{P}(Y | X, \mathcal{M})$.

\begin{algorithm}[!b]
\caption{Interacting agent selection algorithm}\label{alg:alg}
\begin{algorithmic}[1]
\REQUIRE 
Agents' state $ X_t \in \mathbb{R}^{N\times7} $,
Agents' current and future lanes $ L_t \in \mathbb{R}^{N\times2} $,
Selection threshold $ \mathcal{D} $
\ENSURE
Index list $\mathcal{N}$
\STATE Extract position $P_t$ and velocity $V_t$ from $X_t$;
\STATE Initialize min-distance $d_\text{SL}, d_\text{FL}, d_\text{FF}, d_\text{ML} \gets \mathcal{D}$;
\STATE Initialize neighbor index list $\mathcal{N} \gets [0,0,0,0]$;
\FOR{each agent $n \gets 1$ \textbf{to} $N$}
\STATE Distance of n to target $d_t^n \gets \|p_t^{n}-p_t^0\|_2$;
\STATE Orientation of n to target $o_t^{n,0} \gets (p_t^{n}-p_t^0) \cdot v_t^0$;
\STATE Orientation of target to n $o_t^{0,n} \gets (p_t^{0}-p_t^n) \cdot v_t^n$;
\IF{$d_t^n < d_\text{SL} \ and \ l_t^n = l_t^0 \ and \ o_t^{n,0} \ge 0$}
\STATE  $d_\text{SL} \gets d_t^n $;
\STATE  $\mathcal{N}[0] \gets n $;
\ELSIF{$d_t^n < d_\text{FL} \ and \ l_t^n = l_{t+}^0 \ and \ o_t^{n,0} \ge 0 \ and \ o_t^{0,n} < 0 \ and \ l_{t+}^0 \ne l_{t}^0$}
\STATE  $d_\text{FL} \gets d_t^n $;
\STATE  $\mathcal{N}[1] \gets n $;
\ELSIF{$d_t^n < d_\text{FF} \ and \ l_t^n = l_{t+}^0 \ and \ (o_t^{n,0} \ge 0 \ and \ o_t^{0,n} \ge 0 \ or \ o_t^{n,0} < 0)\ and \ l_{t+}^0 \ne l_{t}^0$}
\STATE  $d_\text{FF} \gets d_t^n $;
\STATE  $\mathcal{N}[2] \gets n $;
\ELSIF{$d_t^n < d_\text{ML} \ and \ l_t^0 = l_{t+}^n \ and \ o_t^{n,0} \ge 0 \ and \ l_{t+}^n \ne l_{t}^n$}
\STATE  $d_\text{ML} \gets d_t^n $;
\STATE  $\mathcal{N}[3] \gets n $;
\ENDIF
\ENDFOR
\RETURN $\mathcal{N}$;
\end{algorithmic}
\label{alg}
\end{algorithm}
\subsection{Prediction Model}
We named our proposed model ASPILin, emphasizing agent selection, physical interactions. An overview of our proposed ASPILin is illustrated in 
Fig.~\ref{fig2}.
\subsubsection{Interacting Agent Selection}
We define the Euclidean distance between agent $n$ and the target agent at time $t$ as $d_t^{n}$. If $d_t^{n}$ falls below a manually defined selection threshold $\mathcal{D}$, it is considered that there is a potential for interaction between agent $n$ and the target agent.

In urban areas, vehicles operate within lanes while driving, which allows us to convert the problem of interaction between agents into a correlation problem between lanes. At time $t$, the lane to which agent $n$ belongs is defined as $l_t^{n}$. As time passes, the agent will move to another lane, which we refer to as future lane $l_{t+}^{n}$ ($t+ \leq T_f$). Because the size of each scene is finite, if $l_t^n$ represents the last lane traversed by agent $n$ in the scene, then it is stipulated that $l_{t+}^n=l_{t}^n$. The future lane $l_{t+}^n$ can be easily captured for training data. Consequently, a lane predictor must be built at the inference stage to predict $l_{t+}^n$. Predicting the future lane can easily be converted into a classification problem. However, this method is unsuitable for real-world applications as it only accommodates predictions in scenarios with training. For this purpose, we use an ultra-lightweight model Lin to forecast unimodal medium-to-high-precision trajectories for future moments, then map each timestep of the trajectory onto the respective lane. Lin is a simplified version of ASPILin, which excludes the interaction encoder in Fig.~\ref{fig2}. The entire process can be expressed as $\{l_k^n\}_{k=-T_h+1}^0 = g(\text{Lin}([\{x_k^n\}_{k=-T_h+1}^0,\mathcal{M}]))$, where $g$ is the mapping from trajectory to lane. As an intermediate model, Lin can be substituted with any other trajectory prediction model. However, the choice of model necessitates a trade-off between efficiency and prediction accuracy, a decision that is inherently dependent on the dataset (see Tab.~\ref{tab:table5}).

Then we select four different types of interacting agents as described in Alg.~\ref{alg:alg}: \textit{Same lane and Leading} (SL), \textit{Future lane and Leading} (FL), \textit{Future lane and Following} (FF), and \textit{Merging and Leading} (ML). In brief, the interacting agents SL and ML have already been illustrated in Fig.~\ref{fig1}. As for FF and FL, consider the right blue vehicle in the left subfigure of Fig.~\ref{fig1} as the target vehicle (i.e., the red car). In this case, the original front blue vehicle serves as its FL and the red vehicle becomes its FF.

\subsubsection{Interaction Representation}
As previously stated, applying our agent selection method only to the current time point might lead to the loss of causal links in historical trajectories. Hence, we define the interaction representation as $I = [I^{0}, ( I^{s} )_{s \in \mathcal{S}}] \in \mathbb{R}^{5 \times T_h \times 7}$, where $I^s = [x_{-T_h+1}^s,x_{-T_h+2}^s,\dots,x_{0}^s] \in \mathbb{R}^{T_h \times 7}$ represents the states of type $s$ agents in all observation timesteps and $\mathcal{S} = \{ \text{SL, FL, FF, ML} \}$. Both methods use the same spatial resources. Similar to recent studies\cite{Mo2022MultiAgentTP,Knittel2022DiPAPM,Zhou2022HiVTHV}, we convert the coordinate system for the final interaction representation. Specifically, all states in $I$ are transformed into a relative coordinate system as $\tilde{I}$ with the target agent's final observation point $p_0^0$ as the origin and the positive direction $h_0^0$ of the x-axis. 

\subsubsection{Interaction Encoding}

As mentioned in\cite{Jiang2024InterHubAN}, a crucial aspect of driving interaction is that spatiotemporal conflicts prompt road users to take actions to avoid collisions, inevitably influencing each other's behavior. Thus, we simulate the \textit{original intent} of all agents through a constant acceleration model $\text{CA}(\cdot)$. Then estimate the time $\tau_t^{n}$ needed for the target agent, expressed as:
\begin{equation}
\tau_t^{n} = \underset{\tau}{\arg\min} \|\text{CA}(\tilde{x}_t^0,\tau)-\text{CA}(\tilde{x}_t^{n},\tau)\|_2,
\end{equation}
We set a lower bound for $\tau_t^{n}$ to ensure that the correlation coefficient is based solely on future action and an upper bound based on the assumption that agents will not stay in the scene for more than $T$ seconds. Then the closest distance $d_{t+}^{n}$ can be derived:
\begin{equation}
\overline{\tau}_t^{n} = 
\begin{cases} 
0 & \text{if } \tau_t^{n} < 0; \\
T & \text{if } \tau_t^{n} > T; \\
\tau_t^{n} & \text{otherwise},
\end{cases}
\end{equation}
\begin{equation}
d_{t+}^{n} = \|\text{CA}(\tilde{x}_t^0,\overline{\tau}_t^{n})-\text{CA}(\tilde{x}_t^{n},\overline{\tau}_t^{n})\|_2.
\end{equation}
Intuitively, the closer an agent is to the target agent, the higher the closeness index. However, for a vehicle which is close but moves slowly, another vehicle which is slightly farther but moves faster, may have a greater impact on the target agent. So we construct a closeness index $c_t^{n}$ between agents, which takes into account both the distance and the speed of approach between them, as shown in Eq.~\ref{eq8}. More specifically, we add an extra item $\epsilon$ to the denominator of fraction b to avoid a zero divisor or the generation of an excessively large value that would overshadow the role of fraction a. Similarly, adding $\epsilon$ to the numerator follows the same principle. In the edge case where $d_{t}^{n}=d_{t+}^{n}$ and $\overline{\tau}_t^{n}=0$, $d_{t}^{n}$ will solely dominate $c_t^{n}$.
\begin{equation}
\label{eq8}
c_t^{n} = \underbrace{\frac{1}{d_t^{n}}}_{\textit{a}} \cdot \underbrace{\frac{d_t^{n} - d_{t+}^{n} + \epsilon}{\overline{\tau}_t^{n}+\epsilon}}_{\textit{b}}.
\end{equation}
Next, $c_t^{s} = c_t^{n}$ (agent $n$ belongs to tpye $s$) is normalized to a physical attention score:
\begin{equation}
\alpha_t^{s} = \frac{c_t^s}{\sum_{s \in \mathcal{S}} c_t^s},
\end{equation}
where $\alpha_t^{s}$ represents the spatial relevance of interacting agents to the target agent. The weighted sum of $\alpha$ equals 1 for the same target vehicle. Subsequently, we derive a weight matrix $\mathcal{A} \in \mathbb{R}^{4 \times T_h}$:
\begin{equation}
\mathcal{A} = 
\begin{bmatrix} \alpha_{-T_h+1}^{\text{SL}} & \cdots & \alpha_{0}^{\text{SL}}\\
\alpha_{-T_h+1}^{\text{FL}} & \cdots & \alpha_{0}^{\text{FL}}\\
\alpha_{-T_h+1}^{\text{FF}} & \cdots & \alpha_{0}^{\text{FF}}\\
\alpha_{-T_h+1}^{\text{ML}} & \cdots & \alpha_{0}^{\text{ML}}
\end{bmatrix}
=
\begin{bmatrix} \mathcal{A}^{\text{SL}}\\
\mathcal{A}^{\text{FL}}\\
\mathcal{A}^{\text{FF}}\\
\mathcal{A}^{\text{ML}}
\end{bmatrix}
\end{equation}
Ultimately, $e_{int}$ is obtained by applying residual connection and layer norm:
\begin{equation}
Z = \text{FC}(\tilde{I}) = [Z^0, (Z^{s})_{s \in \mathcal{S}}], 
\end{equation}
\begin{equation}
Z = \text{FC}\left(Z^0 + \sum_{s \in \mathcal{S}} \mathcal{A}^{s} \circ Z^{s}\right),
\end{equation}
\begin{equation}
e_{int} = \text{FFN-LN}(\text{FFN}(\text{LN}(Z)))+Z,
\end{equation}
where $\circ$ denotes the Hadamard product. Inspired by NormFormer\cite{Shleifer2021NormFormerIT}, we additionally use a LN placed after the FNN$(\cdot)$ but before the residual connection referred to as FFN-LN$(\cdot)$, which helps enhance training stability.


\subsubsection{Other Components}
1D Convolutional Neural Networks Conv1D($\cdot$) and Gated Recurrent Units GRU($\cdot$) are used to capture the spatial and temporal dependencies of the target agent's historical states in the relative coordinate system. We use three equivalent but independent spatiotemporal encoders corresponding to decoders to obtain the mean $\mu \in \mathbb{R}^{K \times T_f \times 2}$, variance $\sigma \in \mathbb{R}^{K \times T_f}$, and samples $z \in \mathbb{R}^{K \times T_f \times 2}$ used for reparameterization respectively. 
The corresponding embeddings are labeled as $e_{st}^\mu$, $e_{st}^\sigma$, and $e_{st}^z$.
The map selector from HEAT-I-R\cite{Mo2022MultiAgentTP} is utilized as our map encoder. The entire process is as follows:
\begin{equation}
\mathcal{M}' = \text{MLP}(\text{CNNs}(\mathcal{M})),
\end{equation}
\begin{equation}
e_{map} = \text{Sigmoid}(\text{FC}([\mathcal{M}',\text{FC}(x_0^0)])) \circ \mathcal{M}'.
\end{equation}
The mean and variance are forecasted at first by two different MLP decoder. Next, we use the predicted $\sigma$ for sample prediction:
\begin{equation}
z = \text{MLP}([e_{st}^{z},e_{int},e_{map},\text{MLP}(\sigma)]).
\end{equation}
The predicted trajectory $\hat{Y}$ is ultimately derived from a reparameterization formula:
\begin{equation}
\hat{Y} = \mu + \sigma \times z.
\end{equation}

\subsection{Training Objective}
Since ASPILin does not employ stochastic sampling for reparameterization, the conventional CVAE loss function is no longer applicable. Instead, we adopt the loss function of Leapfrog\cite{Mao2023LeapfrogDM}, which is defined as follows:
\begin{equation}
\mathcal{L} = \mathcal{L}_\text{distance} + \lambda\mathcal{L}_\text{diversity},
\end{equation}
\begin{equation}
\mathcal{L}_\text{distance}= \frac{1}{T_f}{\min}_{k=1}^K \sum_{t=1}^{T_f}\|\hat{y}_t^k-y_t\|_2,
\end{equation}
\begin{equation}
\mathcal{L}_\text{diversity} = \frac{\sum_{k=1}^K\sum_{t=1}^{T_f}\|\hat{y}_t^k-y_t\|_2}{\sigma^2KT_f} + \log\sigma^2.
\end{equation}
The loss term $\mathcal{L}_\text{distance}$ remains the same as the original reconstruction loss and employs a Winner-Takes-All strategy to optimize the closest mode. $\mathcal{L}_\text{diversity}$ with $\lambda = 0.02$ is designed to enhance the diversity of forecasted trajectories. The first component improves the prediction diversity in complex scenarios, while the second acts as a regularization term to prevent excessive variance.
\section{Experiments}
\subsection{Experimental Setup}
\subsubsection{Datasets}
We train and evaluate our model on three popular datasets: the INTERACTION dataset\cite{Zhan2019INTERACTIONDA}, the highD dataset\cite{Krajewski2018TheHD}, and the CitySim dataset\cite{Zheng2022CitySimAD}. INTERACTION contains 398,409 and 107,269 sequences for ASPILin's training and validation and 413,548 and 111,493 sequences for Lin's training and validation. Each sequence is sampled at 10Hz, and the task is to use the past 1 second of sequence data to predict the next 3 seconds. In the case of highD, we split the dataset into training, testing, and validation sequences for 7:1:2 ratio and downsample the trajectories to 5Hz, following the same data processing operation in PiP\cite{Song2020PiPPT}. The prediction task involves using the past 3 seconds to predict the next 5 seconds. For CitySim, we use data from two no-signal scenarios, Intersection B and Roundabout A, with the training (61,185 sequences) and validation (15,164 sequences) split in an 8:2 ratio. The trajectory sampling rate for both scenarios is 30Hz, with the task being to predict the future 6 seconds trajectory based on the past 2 seconds.

\subsubsection{Metrics}
For INTERACTION and CitySim, we forecast future trajectories for $K=6$ modes and evaluate the model's performance using minADE\(_{K}\) and minFDE\(_{K}\). minADE\(_{K}\) represents the minimum average error between the predicted trajectory and the ground truth, while minFDE\(_{K}\) denotes the minimum error of the final trajectory point between the two.
For highD, we predict a deterministic unimodal trajectory and evaluate the model using Root Mean Square Error (RMSE), expressed as: 
\begin{equation}
\text{RMSE} = \sqrt{\frac{1}{NT_f}\sum_{i=1}^{N}\sum_{t=1}^{T_f}\|\hat{y}_{t,i}-y_{t,i}\|_2^2},
\end{equation}
where $N$ represents the total number of samples.

\subsubsection{Implementation Details}

\begin{table}[!t]
\caption{Comparison with models on the INTERACTION dataset\label{tab:table2}}
\centering
\setlength{\tabcolsep}{4pt}
\begin{threeparttable}
\begin{tabularx}{\columnwidth}{lcccc}
\toprule
\multirow{2.5}{*}{Model}&\multicolumn{2}{c}{Val} & \multicolumn{2}{c}{Test} \\
\cmidrule(lr){2-3} \cmidrule(lr){4-5}
& minADE\(_{6}\)$\downarrow$ & minFDE\(_{6}\)$\downarrow$ & minADE\(_{6}\)$\downarrow$ & minFDE\(_{6}\)$\downarrow$ \\
\cmidrule(lr){1-5}
HEAT-I-R\cite{Mo2022MultiAgentTP} *  & 0.19 & 0.66 & - & -\\
ITRA\cite{Scibior2021ImaginingTR}  & 0.17 & 0.49 & - & - \\
GOHOME\cite{Gilles2021GOHOMEGH}  & - & 0.45 & 0.2005 & 0.5988\\
joint-StarNet\cite{Janjos2021StarNetJA}  & 0.13 & 0.38 & - & -\\
DiPA\cite{Knittel2022DiPAPM}  & 0.11 & 0.34 & - & -\\
MB-SS-ASP\cite{Janjos2023BridgingTG}  & 0.10 & 0.30 & 0.1775 & \underline{0.5392}\\
SAN\cite{Janjos2022SANSA}  & 0.10 & 0.29 & - & - \\
GMM-CUAE\cite{Janjos2023ConditionalUA}  & \underline{0.10} & \underline{0.28} & - & -\\
HDGT\cite{Jia2022HDGTHD} & - & - & \textbf{0.1676} & \textbf{0.4776} \\
\cmidrule(lr){1-5}
Lin *  & 0.18 & 0.67 & - & -\\
ASPILin  & \textbf{0.07} & \textbf{0.24} & \underline{0.1703} & 0.5448\\
\bottomrule
\end{tabularx}
\begin{tablenotes}
\small
\item *Model that only performs unimodal prediction.
\end{tablenotes}
\end{threeparttable}
\end{table}

\begin{table}[!t]
\centering
\caption{Comparison with models on the highD test set\label{tab:table22}}
\begin{threeparttable}
\begin{tabular}{lccccc}
\toprule
\multirow{2.5}{*}{Model} & \multicolumn{5}{c}{RMSE$\downarrow$} \\
\cmidrule(r){2-6}
& 1s & 2s & 3s& 4s& 5s\\
\cmidrule(lr){1-6}
CV  & 0.11 & 0.35 & 0.73 & 1.24 & 1.86 \\
MMnTP\cite{Mozaffari2023MultimodalMA}  & 0.19 & 0.38 & 0.62 & 0.95 & 1.39 \\
MHA-LSTM\cite{Messaoud2020AttentionBV}   & 0.06 & 0.09 & 0.24 & 0.59 & 1.18 \\
POVL\cite{Mozaffari2023TrajectoryPW}  & 0.12 & 0.18 & 0.22 & 0.53 & 1.15 \\
iNATran\cite{Chen2022VehicleTP}  & \underline{0.04} & \underline{0.05} & 0.21 & 0.54 & 1.10 \\
VVF-TP\cite{Sormoli2023AND}  & 0.12 & 0.24 & 0.41 & 0.66 & 0.98 \\
BAT\cite{Liao2023BATBH}  & 0.08 & 0.14 & 0.20 & 0.44 & 0.62 \\
HLTP\cite{Liao2024ACT}  & 0.09 & 0.16 & 0.29 & 0.38 & 0.59 \\
\cmidrule(lr){1-6}
Lin  & 0.05 & 0.06 & \underline{0.11} & \underline{0.27} & \underline{0.54}\\
ASPILin  & \textbf{0.03} & \textbf{0.04} & \textbf{0.09} & \textbf{0.22} & \textbf{0.43}\\
\bottomrule
\end{tabular}
\end{threeparttable}
\end{table}

The range threshold $\mathcal{D}$ is set to 30/200/45 meters for INTERACTION/highD/CitySim. The upper bound $T$ is set to 30 and extra item $\epsilon$ is set to 1. In our proposed physics-related method, The dimensions of two FCs are configured as 32 and 256, respectively, and the feed-forward module has a dimension of 256. The Conv1D kernel size is set to 3, the output channels to 32, and the GRU's hidden layer dimension to 256. For the map encoder, we use the same settings as HEAT-I-R\cite{Mo2022MultiAgentTP}, where $\mathcal{M}$ is a $400 \times 250$ grayscale map for each scene. The hidden layers of the three decoders are set to (1024, 1024).

ASPILin (3.5M/2.4M/5.8M parameters for INTERACTION/highD/CitySim) and Lin (2.5M/0.8M/2.9M parameters for INTERACTION/highD/CitySim) are trained on a single RTX-4090. We use AdamW as the optimizer, with a cosine annealing  scheduler\cite{Loshchilov2016SGDRSG}. The initial settings for the learning rate, batch size, and training epochs are 1e-3, 64/128/32 for the INTERACTION/highD/CitySim dataset, and 40, respectively.

\subsection{Comparison with State-of-the-art}

 The results in the INTERACTION validation set shown on Tab.~\ref{tab:table2} indicate that ASPILin achieves SOTA performance. Our method yields competitive results on the test set. Interestingly, the miss rate of ASPILin ranks only 7th (which is not shown on the table) on the leaderboard. This is because the test set includes an additional 30\% out-of-distribution samples, and the robustness of the model is somewhat limited due to our deliberate simplification in map modeling. Moreover, as a lightweight model that does not account for interactions, Lin still achieves the semblable performance as models from the past 2-3 years, demonstrating the feasibility of our agent selection approach. The comparison results on highD are shown in Tab.~\ref{tab:table22}. Similar to some work\cite{Mozaffari2023MultimodalMA,Mozaffari2023TrajectoryPW}, we implemented a Constant Velocity (CV) model as a reference baseline. Interestingly, the performance after 2s exhibited by Lin exceeds that of SOTA methods, attributable to its superior CVAE architecture and loss functions. By comparison, ASPILin significantly reduces prediction error after 3s, highlighting the exceptional performance of our interaction modeling for long-term prediction.

\subsection{Ablation Studies}

\begin{table*}[!t]
\caption{Ablation experiments for each component of the interaction module\label{tab:table3}}
\centering
\begin{tabular}{ccccccccccc}
\toprule
\multirow{2.5}{*}{Variant}&\multicolumn{2}{c}{Agent Selection} & \multicolumn{2}{c}{Select Timestep} & \multicolumn{2}{c}{Interactions Encode} & \multicolumn{2}{c}{INTERACTION} & \multicolumn{2}{c}{CitySim} \\
\cmidrule(r){2-3} \cmidrule(r){4-5} \cmidrule(r){6-7} \cmidrule(r){8-9} \cmidrule(r){10-11}
&four lane-related&four closest&all&current&physical&learned& minADE\(_{6}\)$\downarrow$ & minFDE\(_{6}\)$\downarrow$ & minADE\(_{6}\)$\downarrow$ & minFDE\(_{6}\)$\downarrow$ \\
\cmidrule(lr){1-11}
1&& &  & \checkmark&  &\checkmark &   0.095 & 0.303 & 0.983 & 2.196\\
2&& \checkmark & &\checkmark &  &\checkmark &   0.092 & 0.306 & 0.994 & 2.227\\
3&\checkmark&  & &\checkmark &  &\checkmark &   0.090 & 0.277 & 0.951 & 2.105\\
4&\checkmark&  &\checkmark & &  &\checkmark &   0.088 & 0.278 & 0.935 & 2.059\\
5&\checkmark&  & &\checkmark &\checkmark  & &   0.073 & 0.247 & 0.924 & 2.060\\
\cmidrule(lr){1-11}
6&\checkmark&  &\checkmark & &\checkmark  & &   \textbf{0.069} & \textbf{0.236} & \textbf{0.901} & \textbf{2.024}\\
\bottomrule
\label{tabel3}
\end{tabular}
\end{table*}

\begin{table}[!t]
\caption{Ablation experiments for four types of interacting agents\label{tab:table4}}
\centering
\setlength{\tabcolsep}{4pt}
\begin{tabularx}{\columnwidth}{cccccccc}
\toprule
\multirow{2.5}{*}{SL}&\multirow{2.5}{*}{FL}&\multirow{2.5}{*}{FF}&\multirow{2.5}{*}{ML}&\multicolumn{2}{c}{INTERACTION} & \multicolumn{2}{c}{CitySim} \\
\cmidrule(lr){5-6} \cmidrule(lr){7-8}
&&&& minADE\(_{6}\)$\downarrow$ & minFDE\(_{6}\)$\downarrow$ & minADE\(_{6}\)$\downarrow$ & minFDE\(_{6}\)$\downarrow$ \\
\cmidrule(lr){1-8}
& \checkmark &\checkmark  &\checkmark &   0.073 & 0.251 & 0.973 & 2.147\\
 \checkmark &&\checkmark   &\checkmark &   0.085 & 0.271 & 0.944 & 2.137\\
\checkmark&  \checkmark & &\checkmark &   0.090 & 0.299 & 0.958 & 2.081\\
\checkmark  &\checkmark &\checkmark &&   0.080 & 0.269 & 0.915 & 2.052\\
\cmidrule(lr){1-8}
\checkmark&\checkmark  &\checkmark &\checkmark  &   \textbf{0.069} & \textbf{0.236} & \textbf{0.901} & \textbf{2.024}\\
\bottomrule
\label{tabel4}
\end{tabularx}
\end{table}

\subsubsection{Components of the Interaction Module}

The ablation experiments for each component of the interaction module are shown in Tab~\ref{tab:table3}. The baseline configuration selects all agents within $\mathcal{D}$ at time $t=0$ as interacting agents and encodes their interactions using Transformer. Additionally, we introduce an extra simple agent selection method, which selects the four closest agents to verify the effectiveness of our method. An intuitive conclusion is that merely setting an upper limit on the number of interaction agents does not enhance model performance and may even reduce it. This is reasonable, as inappropriately narrowing the selection range will likely exclude genuinely interacting agents.  From comparisons in variants 2 and 3, we conclude that refining agent selection through lane usage can improve model performance, which provides valuable insights for future research. Switching the time window from current to all enhances model performance, demonstrated across two comparison sets (variants 3 and 4, 5 and 6). The enhancement is particularly evident in the CitySim dataset, owing to the use of extended historical sequences for prediction, thereby elevating the probability of alterations among interacting agents. The last two comparisons (variants 3 and 5, 4 and 6) demonstrate that integrating physical interaction encoding is viable and advantageous, increasing the model's interpretability.

\subsubsection{Four Types of Interacting Agents}

\begin{table}[!t]
\caption{Ablation experiments for different lane predictors\label{tab:table5}}
\centering
\begin{tabular}{lccc}
\toprule
\multirow{2.5}{*}{Model}&\multicolumn{3}{c}{INTERACTION}\\
\cmidrule(lr){2-4}
&ACC(\%)$\uparrow$ & minADE\(_{6}\)$\downarrow$ & minFDE\(_{6}\)$\downarrow$\\
\cmidrule(lr){1-4}
LSTM & 90.1(\textcolor{blue}{\scalebox{0.8}{-9.9\%}}) & 0.077(\textcolor{blue}{\scalebox{0.8}{-18.5\%}}) & 0.251(\textcolor{blue}{\scalebox{0.8}{-8.7\%}}) \\
Lin & 97.9(\textcolor{blue}{\scalebox{0.8}{-2.1\%}}) & 0.069(\textcolor{blue}{\scalebox{0.8}{-6.1\%}}) & 0.236(\textcolor{blue}{\scalebox{0.8}{-2.2\%}})\\
Raw Data & \textbf{100} & \textbf{0.065} & \textbf{0.231}\\
\cmidrule(lr){1-4}
\multirow{2.5}{*}{Model}&\multicolumn{3}{c}{CitySim}\\
\cmidrule(lr){2-4}
&ACC(\%)$\uparrow$ & minADE\(_{6}\)$\downarrow$ & minFDE\(_{6}\)$\downarrow$\\
\cmidrule(lr){1-4}
LSTM & 83.6(\textcolor{blue}{\scalebox{0.8}{-16.4\%}}) & 0.925(\textcolor{blue}{\scalebox{0.8}{-4.5\%}}) & 2.048(\textcolor{blue}{\scalebox{0.8}{-2.2\%}}) \\
Lin& 89.5(\textcolor{blue}{\scalebox{0.8}{-10.5\%}}) & 0.901(\textcolor{blue}{\scalebox{0.8}{-1.8\%}}) & 2.024(\textcolor{blue}{\scalebox{0.8}{-1.0\%}})\\
Raw Data & \textbf{100} & \textbf{0.885} & \textbf{2.003}\\
\bottomrule
\label{tabel4}
\end{tabular}
\end{table}

According to the results in Tab.~\ref{tab:table4}, excluding any category of interacting agents results in some level of decline in model performance. Models excluding FF or FL, achieve the poorest performance on INTERACTION and CitySim, indicating that agents on the target agent's future lane have a more significant impact on the target agent than other agents. Moreover, on the CitySim validation set, the prediction task is more sensitive to changes in interaction agents because of its 6-second prediction horizon.
\begin{figure*}[!t]
\centering
\includegraphics[width=\textwidth]{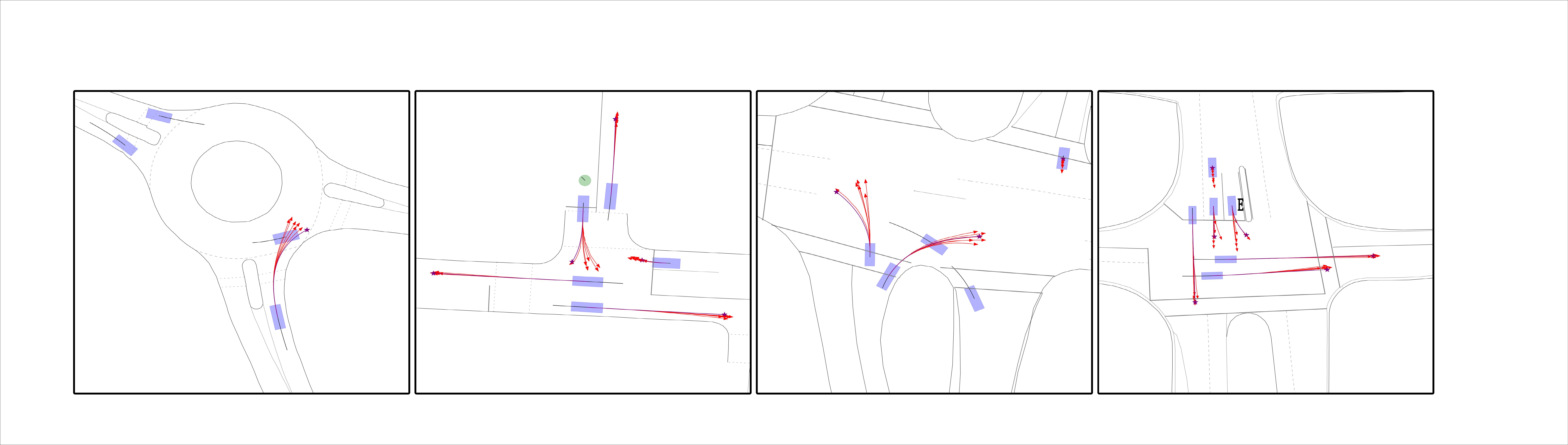}
\caption{Qualitative results on INTERACTION. Purple rectangles represent vehicles, while green circles denote pedestrians or cyclists. Past trajectories are shown with black lines, predicted future trajectories with red lines, and ground truth trajectories with purple lines, with endpoints marked distinctively.}
\label{fig3}
\end{figure*}

\subsubsection{Different Lane Predictors}
We examine how the accuracy of future lane predictions affects model performance by a simple LSTM and Lin. The detailed experimental results are shown in Tab.~\ref{tab:table5}. Using raw data undoubtedly achieves the best performance. It is noteworthy that in interaction-rich datasets like INTERACTION, the differences between lane predictions and ground truth are magnified in the ultimate trajectory predictions, while the inverse scenario is observed in the case of CitySim. This provides valuable insights for the selection of intermediate models: in datasets characterized by interaction scenarios, priority should be given to models with superior performance, whereas in other cases, models with higher efficiency should be prioritized. Moreover, what is not displayed in the table is that Lin's ADE and FDE on the CitySim dataset are 2.809 and 7.785, respectively. Nevertheless, it maintains a high lane prediction accuracy, demonstrating that our comprehensive agent selection strategy is effective even for long-term prediction tasks with a simple model.

\subsubsection{Part of Physical Coefficient Formula}
Through another ablation experiment, we validate the effectiveness of the two components in the physical coefficient formula, with results shown in Tab.~\ref{tab:table7}. While the results show that both \textit{a} and \textit{b} positively influence the predictions, their impact varies notably depending on different datasets. For INTERACTION, the component \textit{b} representing the approach speed has a greater impact on prediction performance. This happens because using pure distance to represent attention between agents leads the model to ignore those that are farther away but have a larger impact. INTERACTION features more complex scenarios, where such agents are more common. In comparison, the prediction task on CitySim, while long-term, involves fewer complex scenarios, which results in the improvements brought by \textit{a} and \textit{b} being nearly equivalent.

\begin{table}[!t]
\caption{Ablation experiments for physical coefficient formula\label{tab:table7}}
\centering
\setlength{\tabcolsep}{4pt}
\begin{tabular}{cccccc}
\toprule
\multicolumn{2}{c}{Part}&\multicolumn{2}{c}{INTERACTION} & \multicolumn{2}{c}{CitySim} \\
\cmidrule(lr){1-2} \cmidrule(lr){3-4} \cmidrule(lr){5-6}
\textit{a}&\textit{b}& minADE\(_{6}\)$\downarrow$ & minFDE\(_{6}\)$\downarrow$ & minADE\(_{6}\)$\downarrow$ & minFDE\(_{6}\)$\downarrow$ \\
\cmidrule(lr){1-6}
 \checkmark & &   0.092 & 0.293 & 0.950 & 2.073\\
& \checkmark &   0.085 & 0.263 & 0.964 & 2.105\\
\cmidrule(lr){1-6}
\checkmark&\checkmark  &   \textbf{0.069} & \textbf{0.236} & \textbf{0.901} & \textbf{2.024}\\
\bottomrule
\label{tabel7}
\end{tabular}
\end{table}

\subsection{Inference Latency}
\begin{table}[!t]
\caption{Inference latency for two datasets\label{tab:table6}}
\centering
\begin{tabular}{llccccc}
\toprule
Dataset & Model & LP & AS & TP & Total(ms)$\downarrow$\\
\midrule
\multirow{2}{*}{INTERACTION} &baseline & - & 22.29 & 0.51 & 22.80 \\
 & ASPILin & 1.00 & \textbf{8.52} & \textbf{0.24} & \textbf{9.76} \\
 \midrule
\multirow{2}{*}{CitySim} &baseline & - & 6.66 & 0.79 & \textbf{7.45} \\
 & ASPILin & 0.99 & \textbf{6.53} & \textbf{0.26} & 7.78 \\
\bottomrule
\end{tabular}
\end{table}
We evaluate the inference latency of the entire prediction process as shown in Tab.~\ref{tab:table6}, including Lane Prediction (LP), Agent Selection (AS), and Trajectory Prediction (TP) on INTERACTION and CitySim and compare it with the baseline (i.e., variant 1 in Tab.~\ref{tab:table3}). AS consumes most computational resources due to extensive data processing and conditional filtering. Even though ASPILin employs more criteria for meticulous agent selection, it remains more efficient than the baseline. In the INTERACTION dataset, up to 25 agents can be within 30 meters of the target agent, while in CitySim, there are a maximum of only 5 agents within 45 meters. This explains the discrepancies between ASPILin and the baseline in the two datasets. Results show that ASPILin possesses high inference efficiency, particularly in interaction-rich scenes. While its efficiency slightly underperforms the baseline in scenarios with low agent density, this is entirely acceptable.

\subsection{Qualitative Results}
We present the qualitative results of ASPILin on the INTERACTION dataset. As illustrated in Fig.~\ref{fig3}, our model can predict accurate multimodal vehicle trajectories in complex scenarios. Nevertheless, not all predictions provided by ASPILin are feasible (e.g., predicting the trajectory of vehicle E moving forward), which indicates potential directions for future improvements.

\section{Conclusion}
In this work, we explore the possibility of interpretable interaction modeling for trajectory prediction from two perspectives: (\romannumeral1) a lane-related method for a more detailed selection of interacting agents, and (\romannumeral2) a physically-related interaction encoding method. We designed a model named ASPILin and conducted experiments on popular datasets. The results indicate that our approach positively affects trajectory prediction, offering substantially increased interpretability over earlier methods. One limitation of this study lies in its assumption that vehicle interactions are inferred exclusively through lane-based criteria, omitting the factor of traffic signals which is applicable to signal-free scenarios. A direction for future research is to propose a broader and more sophisticated approach to agent selection.

\bibliographystyle{IEEEtran}
\bibliography{citation.bib}

\vfill

\end{document}